\documentclass[letterpaper,twocolumn,10pt]{article}
\usepackage{usenix-2020-09}

\usepackage{tikz}
\usepackage{amsmath}
\usepackage{algorithm}
\usepackage{algpseudocode}

\begin{document}
\date{}

\title{\Large \bf Integration of Robotics, Computer Vision, and Algorithm Design: A Chinese Poker Self-Playing Robot}

\author{
{\rm \ \ \ \ \ \ \ \ \ \ \  Kuan-Huang Yu}\\
\and
} 

\maketitle

\begin{abstract}
This paper presents Chinese Poker Self-Playing Robot, an integrated system enabling a TM5-900 robotic arm to independently play the four-person card game Chinese poker. The robot uses a custom sucker mechanism to pick up and play cards. An object detection model based on YOLOv5 is utilized to recognize the suit and number of 13 cards dealt to the robot. A greedy algorithm is developed to divide the 13 cards into optimal hands of 3, 5, and 5 cards to play. Experiments demonstrate that the robot can successfully obtain the cards, identify them using computer vision, strategically select hands to play using the algorithm, and physically play the selected cards in the game. The system showcases effective integration of mechanical design, computer vision, algorithm design, and robotic control to accomplish the complex task of independently playing cards. 
\end{abstract}
\section{Introduction}
Chinese Poker is a popular card game among the Taiwanese elderly, which typically requires four players to play. However, similar to mahjong, it is common to encounter situations where a fourth player cannot be found, resulting in an inability to play the game. In order to address this issue, we proposed a robot that can automatically play Chinese Poker using a sucker mechanism to pick up the cards, while applying YOLOv5 for object detection and classification. YOLOv5 \cite{yolov5}is a classic model for object detection. Its architecture allows the achievement of object detection in single stage with an input RGB image, which making it much faster than models with two stages. This results in more efficient practical application. Moreover, our own greedy algorithm is designed for selecting the cards to be played. This allows the robot to perform all actions from dealing to playing cards.
In order to better understand Chinese Poker, it is necessary to explain the process of game. Four players are each dealt thirteen cards, which they divide into three, five, and five piles and keep face down. The order of the piles is from smallest to largest, with the comparison being made first by card type and then by numerical value \cite{tan2018mastering}. The highest hand is a straight flush, which is five cards of the same suit in sequential order. Next is four of a kind, which is four cards of the same rank plus one other card. The third highest is a full house, which contains three cards of one rank and two cards of another rank. Next is a flush, which is five cards of the same suit but not in sequence. The fifth highest is a straight, which is five cards of sequential ranks but different suits. Then there is three of a kind, containing three cards of the same rank and two other cards. The second lowest hand is two pair, which has two cards of the same rank, another two cards of a different but matching rank, and one extra card. Finally, the lowest possible hand is one pair, which contains two cards of the same rank and three other non-matching cards. This order determines which player wins when comparing the hands in Chinese poker.Once all players have decided on the cards they play, they will reveal cards. Then the sizes of each hands of two people are compared. Every pair of players compares their cards, and if a player's card type is larger than their opponent's, they receive one point, while if it is smaller, they lose one point. If one player's three piles are all larger than their opponent's, the score gained during the comparison will be multiplied by two. This work contributes to gaming robotics and human-robot interaction by showcasing an application involving complex perception and decision making. The techniques could generalize to future assistive agents or game-playing systems.

\section{Related Work}
In recent years, several studies have focused on designing computer programs or robots to play poker as well as board games. AlphaGo, developed by DeepMind, is a prominent example of a computer program that uses artificial intelligence (AI) techniques, particularly deep learning and neural networks to play board games\cite{alphago}. It gained international attention in 2016 when it defeated the world champion Go player, Lee Sedol, in a five-game match. In  \cite{Correia_Alves-Oliveira_Ribeiro_Melo_Paiva_2021},  a social robotic player was introduced for the card game Sueca, addressing challenges in hidden information gaming with the Perfect Information Monte Carlo algorithm. Incorporating emotional intelligence via the FAtiMA framework, the robot achieved a 60\% winning rate in a user study, fostering increased trust levels similar to those observed with human partners.
For robots, computer vision techniques play a crucial role in understanding and interpreting visual information from the world. YOLOv5 is used in various fields for robotic object detection. \cite{LIU2023106217} addressed challenges in robot object detection, proposing a lightweight algorithm based on improved YOLOv5. Introducing C3Ghost, GhostConv, DWConv, and Coordinated Attention modules into YOLOv5 optimizes feature extraction, speeds up detection, and enhances accuracy. In \cite{10090557}, a deep learning-based AIMBOT for game images was proposed, leveraging the YOLOv5 model for fast and accurate detection.The AI vision-based self-targeting system exceled in tracking and shooting tests on the Aimbot game platform, showcasing superior performance compared to competitors. Additionally, strategies used to play poker was investigated in \cite{schaeffer1999learning}, where a the poker program named Loki employed implicit learning by selectively sampling opponent cards and simulating the game, alongside explicit learning involving opponent observation and dynamic play adaptation.
\section{Methodology}
\subsection{System Overview}
\begin{figure}
\begin{center}
\includegraphics{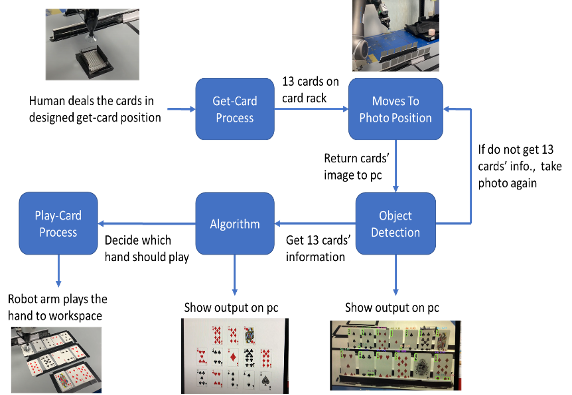}
\end{center}
\caption{\label{fig:system} The overall framework of our system.}
\end{figure}

The overall framework is shown in Figure~\ref{fig:system}. Concretely, three human players will deal 13 poker cards to robot arm in designed get-card position. After that, the robot arm will go to the get-card process to put the 13 cards from designed get-card position to the card rack by our designed sucking process. After 13 cards are all on the card rack, the robot arm will move to photo position. The action is trying to get the cards image to computer. Next, the computer will do the object detection module to get the 13 cards actual suit and number by YOLOv5. If the detection did not get the 13 cards’ information, the robot arm will go back to get the image again. After getting the correct cards’ information, the computer will go to algorithm module to find the best set of hands. The object detection module and the algorithm module’s result will show on the computer to double check. Lastly, the robot arm will do the play-card process to play the hands in order and finish the entire round of Chinese poker card game. In the next section, each module will be elaborated in more detail.

\subsection{System Mechanism}

\begin{figure}
\begin{center}
\includegraphics{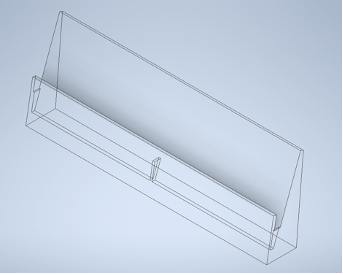}
\end{center}
\caption{\label{fig:cad} The CAD design of card rack.}
\end{figure}

\begin{figure}
\begin{center}
\includegraphics{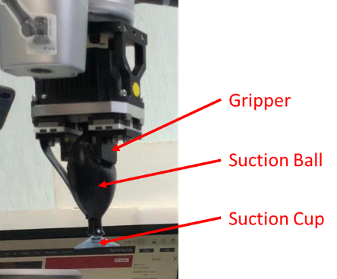}
\end{center}
\caption{\label{fig:sucker} The sucker mechanism.}
\end{figure}

\paragraph{Hardware design}
There are three hardwares we need to design. The first one is the card rack on the get-card position. We just use a box fit in one card size. By using this, when the robot arm sucks more than one card and trys to get rid of others cards, the other cards will stay in a fixed potion and it is easy for get the second card because the position is same as last. 
The second one is the card rack to let the robot arm put the poker cards. The CAD design of this card rack is show in Figure~\ref{fig:cad}. We design some slope and slot to ensure that cards are in a stable position. And there are a small board to block the suit and number of the botton of poker card. The reason is that we want to make sure that the photo will not get these information to confuse YOLOv5 object detection.
The last hardware we designed is sucker mechanism. The sucker mechanism is illustrated in Figure~\ref{fig:sucker}, we use a suction ball to generate the suction force and use a suction cup to touch poker cards. In this design, when the robot arm’s gripper grips, it will squeeze the suction ball and then if the suction cup touch card, gripper releases suction ball to make some negative pressure in the ball. Since poker card will not leak air from suction ball and also poker card has very light weight, the suction force is strong enough to suck one card. Therefore, after doing the process above, poker card will be suck in tightly. For the releasing part, the gripper will grip and  release once to squirt the card out of the suction cup. 

\paragraph{Transformation from camera to robot base}
In order to suck the cards precisely, we have to transform the coordinate of a given point in camera frame into the position of the robot arm’s end effector. This process is called frame transformation which derive the transformation matrix ${ }^{\text {base }} T_{\text {camera }}$ as follows \cite{shih2018simple}. Let $C=\left[c_x, c_y, 1\right]^T$ be a 3-dimentional vector where $c_x$ and $c_y$ represent $x$ axis and $y$ axis of the card in camera frame coordinate respectively. And $B=\left[b_x, b_y, 1\right]^T$ be a 3-dimensional vector where $b_x$ and $b_y$ represent $x$ axis and $y$ axis of the card in base frame coordinate respectively. First, we have to calculate the relationship of $B$ and $C$ vector. 
\begin{equation}
\label{eqn:1}
B={ }^{\text {base }} T_{\text {camera }} \times C
\end{equation}
We will determine three different points $b_1=\left[b_{1 x}, b_{1 y}, 1\right]^T$, $b_2=\left[b_{2 x}, b_{2 y}, 1\right]^T$, $b_3=\left[b_{3 x}, b_{3 y}, 1\right]^T$ in robot base frame coordinate from the workspace. And, we move robot arm to the predefine position to take a picture for workspace. After that, we can get $b_1$,$b_2$,$b_3$ points’ pixel value in the camera frame coordinate respectively and denote as $c_1=\left[c_{1 x}, c_{1 y}, 1\right]^T$, $ c_2=\left[c_{2 x}, c_{2 y}, 1\right]^T$, $c_3=\left[c_{3 x}, c_{3 y}, 1\right]^T$.
Due to the unit from camera and robot base coordinate are different, we have to calculate r which is the ratio of pixel to mm. Then, we take three different points with camera frame coordinate and robot base frame coordinate, it is easy to get the transformation matrix by inverse the camera frame matrix $C$. And the \eqref{eqn:1} can be rewritten as: 
\begin{equation}
\label{eqn:2}
{ }^{\text {base }} T_{\text {camera }}=B \cdot C^{-1}
\end{equation}
, where $$
B=\left[\begin{array}{ccc}
b_{1 x} & b_{2 x} & b_{3 x} \\
b_{1 y} & b_{2 y} & b_{3 y} \\
1 & 1 & 1
\end{array}\right], C=\left[\begin{array}{ccc}
C_{1 x} & C_{2 x} & C_{3 x} \\
C_{1 y} & C_{2 y} & C_{3 y} \\
1 & 1 & 1
\end{array}\right]
$$
After we defined the transformation matrix, we can use this to control the end effector and the workspace’s relationship by using Python to give the moving command to robot arm to get desired point.

\paragraph{Get-card process}
is a series of robot arm process. First, the gripper will squeeze the suction ball to prepare to suck the poker card. Then, the robot arm will go to the predefined get-card position and touch the top card. After that, gripper will release the suction ball to suck the card. And robot arm will shake the gripper to ensure that only sucks one card. Lastly, the robot arm sucks the card to the card rack and grips to release card. In above robot arm’s action, we all use trial and error to let the trajectory smooth by defined middle position.
	
\paragraph{Move to photo position}
 is a process to take a photo of 13 poker cards’ image. After doing the Get-card process 13 times, the robot will get all the card it need to get. And then, the robot arm will do this Move to photo position process. To ensure one image will get all cards’ information and the cards in the image will be perfect rectangular, the postion and orientation of robot arm’s camera on the base frame is defined as $(x, y, z, \alpha, \beta, \gamma)=(100,610,250,125,0,45)$.

\paragraph{Play-card process}
is a series of robot arm process. After the algorithm decide the card hands order, the robot arm will do this Play-card process. In this process, the robot arm will play the hands in order of 3 cards, 5 cards, 5cards by algorithm. The detail will elaborate following. First, the gripper will squeeze the suction ball to prepare to suck the poker card. Then, the robot arm will vertically touches the card which needs to play on the card rack and releases the suction ball to suck. Lastly, robot arm will suck the card, move to the desired play-card position and release the card. After doing this process 13 times, the robot arm finish the entire round of Chinese poker card game. In above robot arm’s action, we all use trial and error to let the trajectory smooth by defined middle position. 

\subsection{Object Detection}
\begin{figure}
\begin{center}
\includegraphics{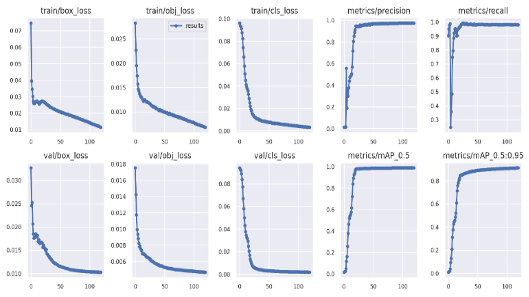}
\end{center}
\caption{\label{fig:yolo} Yolov5 Training result}
\end{figure}
\begin{figure}
\begin{center}
\includegraphics{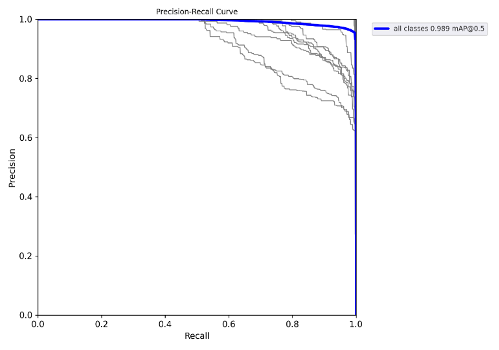}
\end{center}
\caption{\label{fig:valid} Precision-recall Curve of validation dataset}
\end{figure}
\begin{figure}
\begin{center}
\includegraphics{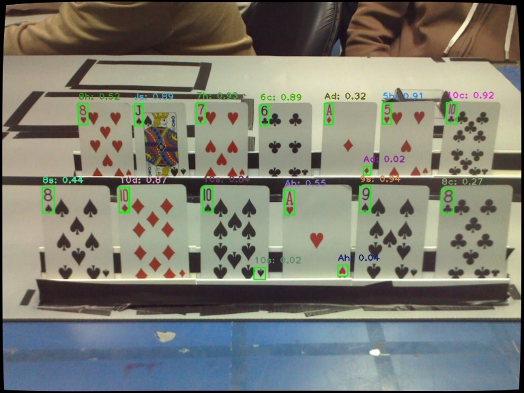}
\end{center}
\caption{\label{fig:img} The detection result of all 13 cards, including the predicted bounding box and the card suit and number}
\end{figure}

 After the robot takes the photo of all 13 cards, we need to determine the number and suit of each card. To handle this case, we obtain object detection technology by introducing deep learning model to predict the card. For the dataset, we downloaded the poker card dataset from roboflow\cite{playing-cards-ow27d_dataset}, and collected about 14000 images for training, 4000 images for validation, and 4000 images for testing, with 52 classes in total representing 52 kinds of poker cards. All images include one to several poker cards with various types of backgrounds, and training and validation images include classes and bounding box coordinate of each class. We chose YOLOv5 as our final training model, resized the image into 416$\times$416, set batch size to 16, and trained 120 epochs from scratch, which took about one day on GTX 1080 GPU. As the training result, Figure~\ref{fig:yolo} showed that the model performed well on both training and validation dataset with such a low loss (about 0.01) and high precision (near 0.99). In addition, from the precision-recall curve shown in Figure~\ref{fig:valid} of validation dataset, the model achieved 0.989 mAP with 0.5 confidence. We had also tested with our own poker card images taken by our cellphone and confirmed that the model could output correct classes and bounding box positions with different background and different angle of cards, thus showing that the model is robust enough to apply on out task – detect the image taken by robot arm.
Then we introduce the trained model to our system to let the robot inference the photo taken by itself. One detection result example is shown as Figure~\ref{fig:img}, including the output bounding box and the card suit and number. Here we set the confidence to 0.1, and the threshold of non-maximum suppression to 0.45. Because the two card racks contain 13 cards, so we write additional codes to check whether the detection successfully detect all the cards or not. First, owing to there may be 2 or more bounding boxes with different classes that are drawn on the same card, we choose the one with higher confidence in this case. Second, if the output contains two identical classes but the bounding box position is different, we choose the one with higher confidence as well. Third, we filter the cards bounding boxes with the range of the center of the output bounding boxes’ y-coordinate, this is used to filter out the bounding box that is not located at the actual number and suit position of the card. After doing the above steps, if the total output contains is still not equal to 13 classes, we let the robot arm back to photo position and re-take a photo, then do the above three mentioned steps again. Finally, we can get the accurate card information.

\subsection{Divide Hands Algorithm}
The algorithm of dividing the 13 cards the robot has into 3 hands is following the greedy algorithm \cite{Jungnickel1999}  that searches from the largest poker hands to the smallest and the pseudo code is shown in Algorithm 1. Some considerations are taken in this algorithm; for example, if the cards are composed of a straight, four K’s and two pairs, the algorithm will separate the smaller pair for an arbitrary card in four-of-a-kind; for another example, if the cards are composed of six cards in the same suit, a full-house and a pair, the flush will take the largest card in that suit and attach four cards in that suit from small to large; for another example, if the cards are composed of four pairs and a straight-flush, the algorithm will make two pairs by taking the largest and the smallest pairs in the four pairs in order to have the last three cards be greatest; for another example, if the cards are composed of three cards have the same value and ten cards of nothing, the three-of-a-kind will have two smallest cards in the ten cards with it, and the high cards will be made up of a largest card and four smallest cards from the eight remaining cards; least but not the last, for another example, if the cards are composed of three K’s, three Q’s, three J’s and two pairs, then the algorithm will not break the J’s to serve the pairs for the full houses of three K’s and three Q’s to make sure the last three cards are as large as we can.
Besides, before running the algorithm, we have also checked the integrity of the given input, including invalid number of cards, invalid suit or value of a card and duplicated cards to ensure the output is correct. And we also record the position for each card that once the hands are decided, we can use our sucker to suck cards up according to the positions given.

\begin{algorithm}

\caption{divideHandsAlgorithm}

\textbf{Input}: cards\\
\textbf{Output}: divided\_cards\\
\textbf{Initialize }spade, heart, diamond, club $=[\overbrace{\text { False, } \cdots, \text { False }}^{13 \text { times }}$, hands $=[\ ]$, four\_of\_a\_kind, three\_of\_a\_kind, pairs $=[\ ]$.\\
\begin{algorithmic}[1]
      \For{\texttt{card \textbf{in} cards}}
        \State suit, value $=$ card $//$ suit $\in\{$ spade, heart, diamond, club $\}$, value -1 is index
        \State suit $[$ value -1$]=$ True $//$ set it to true because robot has this card
      \EndFor

\State Find cards of the same kind for arbitrary cards in four\_of\_a\_kind, full-houses, etc.

\State Find straight-flushes in cards, put the found cards into divided\_cards, reset the found cards to False, append the last three cards and return divided\_cards if two hands are found. // handle straight-flushes

\State Find four\_of\_a\_kind in cards, put the found cards into divided\_cards,
reset the found cards to False, append the last three cards and return
divided\_cards if two hands are found. // handle four\_of\_a\_kind
    \State Find full-houses in cards, put the found cards into divided\_cards,
reset the found cards to False, append the last three cards and return
divided\_cards if two hands are found. // handle full-houses
    \State Find flushes in cards, put the found cards into divided\_cards,
reset the found cards to False, append the last three cards and return
divided\_cards if two hands are found. // handle flushes
    \State Find straights in cards, put the found cards into divided\_cards,
reset the found cards to False, append the last three cards and return
divided\_cards if two hands are found. // handle straights
    \State Find three\_of\_a\_kind in cards, put the found cards into divided\_cards,
reset the found cards to False, append the last three cards and return
divided\_cards if two hands are found. // handle three\_of\_a\_kind
    \State Find two-pairses in cards, put the found cards into divided\_cards,
reset the found cards to False, append the last three cards and return
divided\_cards if two hands are found. // handle two-pairses
    \State Find one-pairs in cards, put the found cards into divided\_cards,
reset the found cards to False, append the last three cards and return
divided\_cards if two hands are found. // handle one-pairs
    \State Make high-cards in cards, put the cards into divided\_cards, append the last three cards and return divided\_cards. // handle high-cards
 \end{algorithmic}      
\end{algorithm}
\section{Experiment}

\subsection{Environment Setup}
\begin{figure}
\begin{center}
\includegraphics{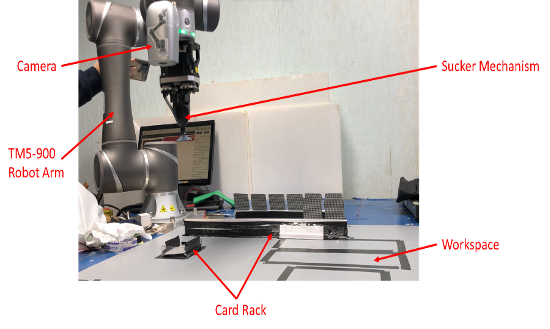}
\end{center}
\caption{\label{fig:setup} The experimental setup.}
\end{figure}

The experimental setup is illustrated in Figure~\ref{fig:setup}, including robot arm, eye in hand camera module, sucker mechanism, card rack and workspace.

\paragraph{Robot Arm}
We use the TM5-900 robot arm, its payload is 4 kilograms with a 900-millimeter reach, the typical speed is 1.4(m/s), and 0.05-millimeter repeatability (repeatable accuracy) \cite{robot}. The robot arm’s workspace and speed are all conformed to our experiment.

\paragraph{Eye in Hand Camera Module}
On the TM5-900 robot arm, there is an eye in hand camera module. We use this camera to get poker cards’ image and then get the cards’ information by object detection.

\paragraph{Sucker Mechanism}
We use a suction ball and a suction cup to build the sucker. The sucker can help us get poker cards and play the cards.

\paragraph{Card Rack}
There are two kinds of card rack in the system. The first kind of rack is on the get-card position. Its function is when robot arm sucks one card and let other cards be a fixed position. The second kind of rack is design for put the cards which robot arm got and it is made by 3D printing.

\paragraph{Workspace}
A space where cards can be played and robot arm can reach. The space divided into three parts, including get-card position, put-cards’ card rack, play-card position.

\subsection{Result}
\begin{figure}
\begin{center}
\includegraphics{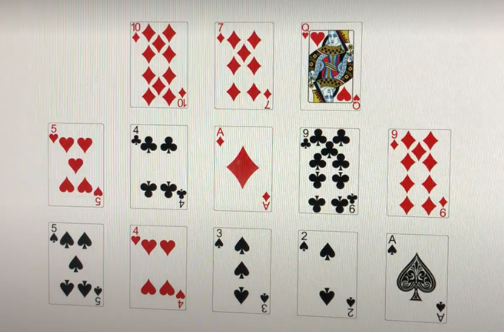}
\end{center}
\caption{\label{fig:exhibit} Cards selection exhibition}
\end{figure}
\begin{figure}
\begin{center}
\includegraphics{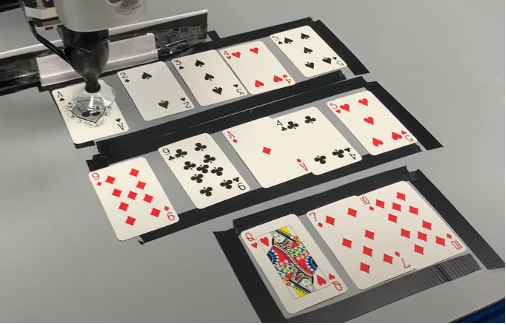}
\end{center}
\caption{\label{fig:result} Experiment result}
\end{figure}

Our designed robot  is capable of effectively completing a series of actions including picking up cards, placing them on the card rack, recognizing the numerical and suit of the cards, selecting the card type, and playing the cards. In order to showcase the playing more effectively, we have also generated a display of the cards selected by the robot \cite{	vector-playing-cards}. Shown in Figure~\ref{fig:exhibit}, the robot selects straight, two pairs, and high card to play. After doing all above process, the Chinese poker self-playing robot playing result is shown in Figure~\ref{fig:result} which is same asFigure~\ref{fig:exhibit} shown. 
Due to the lack of a specific metric for evaluating the optimality of algorithmic combinations for assembling cards, it is challenging to determine if the algorithm has achieved the best solution. However, visual inspection of the images demonstrates that the algorithm, when integrated with a robotic arm, is capable of assembling card formations successfully.

\section{Conclusion}
In this paper, we present a robotic system called Chinese Poker Self-Playing Robot. The robot is capable of playing the card game itself. With the sucker mechanism, the robot can suck up the cards from initial card pile and put them on card racks, or play the cards from the card racks. With YOLOv5 object detection model, the robot can get the correct card information by simply taking a photo of the card racks. With the divide hands algorithm, the robot can determine which card to play. The experiment results show that the robot can get the desired card order and play Chinese Poker card game smoothly. For future work, we can further deal the cards in arbitrary place in the beginning and improve our play card algorithm rather than   greedy algorithm to make the system more robustness. 


\begin{thebibliography}{10}

\bibitem{yolov5}
Ultralytics.
\newblock {ultralytics/yolov5: v7.0 - YOLOv5 SOTA Realtime Instance Segmentation}.
\newblock \url{https://github.com/ultralytics/yolov5.com}, 2022.
\newblock Accessed: 7th May, 2023.

\bibitem{tan2018mastering}
Andrew Tan and Jarry Xiao.
\newblock Mastering open-face chinese poker by self-play reinforcement learning, 2018.

\bibitem{alphago}
David Silver, Aja Huang, Christopher~J. Maddison, Arthur Guez, Laurent Sifre, George van~den Driessche, Julian Schrittwieser, Ioannis Antonoglou, Veda Panneershelvam, Marc Lanctot, Sander Dieleman, Dominik Grewe, John Nham, Nal Kalchbrenner, Ilya Sutskever, Timothy Lillicrap, Madeleine Leach, Koray Kavukcuoglu, Thore Graepel, and Demis Hassabis.
\newblock Mastering the game of go with deep neural networks and tree search.
\newblock {\em Nature}, 529:484--503, 2016.

\bibitem{Correia_Alves-Oliveira_Ribeiro_Melo_Paiva_2021}
Filipa Correia, Patrícia Alves-Oliveira, Tiago Ribeiro, Francisco Melo, and Ana Paiva.
\newblock A social robot as a card game player.
\newblock {\em Proceedings of the AAAI Conference on Artificial Intelligence and Interactive Digital Entertainment}, 13(1):23--29, Jun. 2021.

\bibitem{LIU2023106217}
Gang Liu, Yanxin Hu, Zhiyu Chen, Jianwei Guo, and Peng Ni.
\newblock Lightweight object detection algorithm for robots with improved yolov5.
\newblock {\em Engineering Applications of Artificial Intelligence}, 123:106217, 2023.

\bibitem{10090557}
Yongqi Cui, Macheng Si, and Qi~Li.
\newblock Game image detection and application based on improved yolov5.
\newblock In {\em 2023 IEEE 2nd International Conference on Electrical Engineering, Big Data and Algorithms (EEBDA)}, pages 1012--1017, 2023.

\bibitem{schaeffer1999learning}
Jonathan Schaeffer, Darse Billings, Lourdes Pe{\~n}a, and Duane Szafron.
\newblock Learning to play strong poker.
\newblock In {\em The International Conference on Machine Learning Workshop on Game Playing}, volume~4, 1999.

\bibitem{shih2018simple}
Ching-Long Shih and Yi~Lee.
\newblock A simple robotic eye-in-hand camera positioning and alignment control method based on parallelogram features.
\newblock {\em Robotics}, 7(2):31, 2018.

\bibitem{playing-cards-ow27d_dataset}
Augmented Startups.
\newblock Playing cards dataset.
\newblock \url{ https://universe.roboflow.com/augmented-startups/playing-cards-ow27d }, aug 2023.
\newblock visited on 2023-11-26.

\bibitem{Jungnickel1999}
Dieter Jungnickel.
\newblock {\em The Greedy Algorithm}, pages 129--153.
\newblock Springer Berlin Heidelberg, Berlin, Heidelberg, 1999.

\bibitem{robot}
TECHMAN~ROBOT INC.
\newblock Tm5 guide book, 2017.

\bibitem{vector-playing-cards}
Google Code.
\newblock {vector-playing-cards}.
\newblock \url{https://code.google.com/archive/p/vector-playing-cards/}, 2011.
\newblock [Online; accessed 26-November-2023].

\end{thebibliography}

\end{document}